\pdfoutput=1
\documentclass[11pt,a4paper]{article}
\usepackage[hyperindex,breaklinks]{hyperref}
\usepackage{emnlp2018} 
\usepackage{times}
\usepackage{latexsym}

\usepackage{url}

\usepackage{enumitem}
\usepackage{amsmath}
\usepackage{amssymb}
\usepackage{bbm}
\usepackage{multirow}
\usepackage{booktabs}
\usepackage{multirow}
\usepackage{graphicx}
\usepackage{caption}
\usepackage{comment}
\usepackage{kotex}
\usepackage{changepage}
\usepackage{blindtext}
\usepackage{caption}
\usepackage{ragged2e}
\usepackage{makecell}
\usepackage{tabularx}

\usepackage{xspace}
\usepackage{xifthen}

\newcommand{\DELETE}{\textsc{delete}\xspace}
\newcommand{\COPY}{\textsc{copy}\xspace}

\newcommand{\INSERT}[1][]{%
\ifthenelse{\equal{#1}{}}{\textsc{insert}\xspace}%
{\textsc{insert}$(#1)$}%
}
\newcommand{\SUBST}[1][]{%
\ifthenelse{\equal{#1}{}}{\textsc{subst}\xspace}%
{\textsc{subst}$(#1)$}%
}
\newcommand{\bv}[1]{\mathbf{#1}}

\newcommand{\Loss}{\mathcal{L}}
\newcommand{\loss}{\ell}

\aclfinalcopy 
\def\aclpaperid{1161} 

\usepackage{array}
\newcolumntype{L}[1]{>{\raggedright\let\newline\\\arraybackslash\hspace{0pt}}m{#1}}
\newcolumntype{C}[1]{>{\centering\let\newline\\\arraybackslash\hspace{0pt}}m{#1}}
\newcolumntype{R}[1]{>{\raggedleft\let\newline\\\arraybackslash\hspace{0pt}}m{#1}}

\title{Imitation Learning for Neural Morphological String Transduction}

\author{Peter Makarov\qquad\qquad\qquad\qquad Simon Clematide\\
  Institute of Computational Linguistics \\
  University of Zurich, Switzerland \\
  \qquad{\tt makarov@cl.uzh.ch}\qquad\quad   {\tt simon.clematide@cl.uzh.ch} \\}

\date{}

\begin{document}
\maketitle
\begin{abstract}
We employ imitation learning to train a neural transition-based string
transducer for morphological tasks such as inflection                           
generation and lemmatization. Previous approaches to training this type of model
either rely on an external character aligner for the production of gold
action sequences, which results in a suboptimal model due to the                
unwarranted dependence on a single gold action sequence despite                 
spurious ambiguity, or require warm starting with an MLE model. Our approach
only requires a simple expert policy, eliminating                               
the need for a character aligner or warm start. It also addresses familiar MLE
training biases and leads to strong and state-of-the-art performance on several
benchmarks.
\end{abstract}

\section{Introduction}


Recently, morphological tasks such as inflection generation and lemmatization
(Figure~\ref{fig:morphexample}) have been successfully tackled with neural 
transition-based models over edit actions \cite{Aharoni&Goldberg2017,Robertson&Goldwater2018,Makarov&Clematide2018,Cotterelletal2017a}.
The model, introduced in \newcite{Aharoni&Goldberg2017}, uses                   
familiar inductive biases about morphological string transduction
such as conditioning on a single input character and monotonic
character-to-character alignment. Due to this, the model achieves lower time
complexity (compared to soft-attentional seq2seq models)
and strong performance on several datasets. 

\citeauthor{Aharoni&Goldberg2017} train the model by maximizing the conditional
log-likelihood (MLE) of gold edit actions derived by an independent
character-pair aligner. The MLE training procedure is therefore a pipeline, and
the aligner is completely uninformed of the end task. This results in
error propagation and the unwarranted dependence of the transducer on a single
gold action sequence---in contrast to weighted finite-state transducers
(WFST) that take into account all permitted action sequences. 
Although these problems---as well as the exposure bias and the
loss-metric mismatch arising from this MLE training \cite{Wiseman&Rush2016}---%
can be addressed by reinforcement learning-style methods 
\cite[RL]{Ranzato16,Bahdanauetal2017,Shenetal2016}, 
for an effective performance, all these approaches
require warm-start initialization with an MLE-pretrained model.
Another shortcoming of the RL-style methods is delayed punishment: For many NLP                        
problems, including morphological string transduction, one can pinpoint 
actions that adversely affect the global score. For example, it is easy to tell
if inserting some character $c$ at step $t$ would render the entire output incorrect.
Assigning individual blame to single actions directly---as opposed to
scoring the entire sequence via a sequence-level objective---simplifies the
learning problem.

\begin{figure}[t]
\begin{adjustwidth}{-0.25cm}{}
\small
\centering
\resizebox{1.02\linewidth}{!}{
\begin{tabular}{ccc}
하다  & \multirow{2}{*}{\scalebox{1.5}{$\Huge\rhd$}} & \multirow{2}{*}{하셨습니까} \\  
{\small$\{$\textsc{V, Past, Formal Polite,}} & & \\
{\small\textsc{Interrogative, Honorific}$\}$} & & \\
& & \\
하셨습니까  & \multirow{1}{*}{\scalebox{1.5}{$\Huge\rhd$}} & \multirow{1}{*}{하다} \\  
\end{tabular}}
\end{adjustwidth}
\caption{Morphological tasks with examples in Korean: inflection generation (top) 
and lemmatization (bottom). (McCune-Reischauer: 하다=hada, 하셨습니까=hasy\u{o}ss\u{u}mnikka).}
\label{fig:morphexample}
\vspace{-5mm}
\end{figure}

Faced with problems similar to those arising in transition-based dependency     
parsing with static oracles \cite{Goldberg&Nivre2012}, we train this model
in the imitation learning (IL) framework \cite{Daumeetal2009,Rossetal2011,Changetal2015},  
using a simple expert policy.
Our approach eliminates both all dependency on an external
character aligner and the need for MLE pre-training.                            
By making use of exploration of past and future actions and having a
global objective, it addresses the MLE training biases, while providing
relevant action-level training signal.
The approach leads to strong and state-of-the-art results on a number of
morphological datasets, outperforming models trained with minimum risk training (MRT).


\medskip
\section{Model Description}
\label{sec:model}
We use a variant of the seq2seq state-transition system 
by \newcite{Aharoni&Goldberg2017}. The model transduces the input string into
the output string by performing single-character edits (insertions, deletions).
The encoder RNN computes context-enriched representations of input
characters, which are pushed onto the buffer at the beginning of transduction.
The decoder RNN keeps track of the history of edits. Transitions---edits---are
scored based on the output of the decoder state and can write a character or pop 
the representation of a character from the top of the buffer. We choose the model variant of
\newcite{Makarov&Clematide2018}, who add the copy edit
, which results in strong performance gains in low-resource settings.

Let $\bv{x}=x_1\dots x_n$, $x_i \in \Sigma_x$ be an input \mbox{sequence,} $\bv{y}=y_1\dots y_p$, $y_j \in \Sigma_y$
an output sequence, and $\bv{a}=a_1\dots a_m$, $a_t \in \Sigma_a$ an action sequence.
Let $\{f_h\}^H_{h=1}$ be the set of all features. 
The morpho-syntactic description of a transduction is then an $n$-hot vector $\bv{e} \in \{0, 1\}^H$.

The model employs a bidirectional long short-term memory (LSTM) encoder
\cite{Graves&Schmidhuber2005} to produce representations for each character of the input $\bv{x}$:%
\vspace{-5mm}
\begin{adjustwidth}{-0.25cm}{}%
\begin{equation}
\bv{h}_1,\dots,\bv{h}_n = \textrm{BiLSTM}(E(x_1),\dots,E(x_n)),
\end{equation}%
\end{adjustwidth}%
\vspace{-1mm}
where $E$ returns the embedding for $x_i$. We push $\bv{h}_1,\dots,\bv{h}_n$ in
reversed order onto the buffer. The transduction begins with the full buffer
and the empty decoder state.

Transitions are scored based on the output of the LSTM decoder state \cite{Hochreiter&schmidhuber1997}:
\begin{equation}
\bv{s}_t = \textrm{LSTM}(\bv{c}_{t-1}, [A(a_{t-1}) \;;\; \bv{h}_i ]),
\end{equation}
where $\bv{c}_{t-1}$ is the previous decoder state, $A(a_{t-1})$ is the embedding of
the previous edit action, and $\bv{h}_i$ is the input character representation
at the top of the buffer.                 
If features are part of the                                                     
input in the task, then the input to the decoder also contains the
representation of morpho-syntactic description $\bv{e}$, $[F(f_1) \,; \dots \,; F(f_H)]$,
which is a concatenation of the embedded features and a designated embedding
$F(0)$ is used instead of $F(f_h)$ if $e_h = 0$.  

The probabilities of transitions are computed with a softmax classifier:%
\vspace{-6mm}
\begin{adjustwidth}{-0.25cm}{}%
\begin{equation}
P(a_t = k \mid \bv{a}_{<t}, \bv{x}, \Theta) = \textrm{softmax}_k(\bv{W}\cdot\bv{s}_t + \bv{b})
\label{gen}
\end{equation}%
\end{adjustwidth}%
\vspace{-2mm}
Model parameters $\Theta$ include  $\bv{W}$,         
$\bv{b}$, the embeddings, and the parameters of the LSTMs.  

The alphabet of edit actions $\Sigma_a$ contains \INSERT[c] for each
$c \in \Sigma_y$, \DELETE, and \COPY. An \INSERT[c] action outputs $c$;
\DELETE pops $\bv{h}_i$ from the top of the buffer; \COPY pops
$\bv{h}_i$ from the top of the buffer and outputs $x_i$.
The system exhibits spurious ambiguity: Multiple action sequences lead to the same
output string. 

\subsection{MLE Training}
\citeauthor{Aharoni&Goldberg2017} train their model by minimizing the negative
conditional log-likelihood of the data
$D = \{(\bv{x}^{(l)}, \bv{a}^{(l)})\}_{l=1}^N$:
\vspace{-5mm}
\begin{adjustwidth}{-0.3cm}{}%
\begin{equation}
\Loss(D, \Theta) = - \sum^N_{l=1} \sum^m_{t=1} \log \! P(a^{(l)}_t\! \mid \! \bv{a}^{(l)}_{<t}, \bv{x}^{(l)},\! \Theta),
\end{equation}%
\end{adjustwidth}%
\vspace{-2mm}
where gold action sequences $\bv{a}^{(l)}$ are deterministically computed from
a character-pair alignment of the input and output sequences
$(\bv{x}^{(l)}, \bv{y}^{(l)})$. The character-pair aligner is trained separately
to optimize the likelihood of the actual training data
$T = \{(\bv{x}^{(l)}, \bv{y}^{(l)})\}_{l=1}^N$.
For the details, we refer the reader to \newcite{Aharoni&Goldberg2017}.         

\section{IL Training}
One problem with the MLE approach is that the aligner is trained in a disconnect
from the end task. As a result, alignment errors lead to the
learning of a suboptimal transducer. Switching to a different aligner
can dramatically improve performance \cite{Makarov&Clematide2018}. More fundamentally,
in the face of the vast spurious ambiguity, the transducer is forced to
adhere to a single gold action sequence whereas typically, legitimate and
equally likely alternative edit sequences exist. This uncertainty is not
accessible to the transducer, but could be profitably leveraged by it.

We address this problem within the IL framework and train the model to imitate  
an expert policy (dynamic oracle), which is a map---on the training data---from
configurations to sets of optimal actions. Actions are optimal if they lead to
the lowest sequence-level loss, under the                                        
assumption that all future actions are also optimal \cite{Daumeetal2009}.
%
%
In the \emph{roll-in} stage, we run the model on a training sample and follow
actions either returned by the expert policy (as in teacher forcing) or sampled 
from the model (which itself is a stochastic policy). In this way, we
obtain a sequence of configurations summarized as decoder outputs
$\bv{s}_1, \dots, \bv{s}_m$. In the \emph{roll-out} stage, we compute the
sequence-level loss for every valid action $a$ in each configuration $\bv{s}_t$.
To this end, we execute $a$ and then either query the expert to obtain the loss
for the optimal action sequence following $a$ or run the model for the rest of  
the input and evaluate the loss of the resulting action sequence. Finally, the
sequence-level losses obtained in this way for all actions $a$ enter the
action-level loss for configuration $\bv{s}_t$ that we minimize with respect to $\Theta$.

\paragraph{Sequence-level loss} We define the loss in terms of the
\emph{Levenshtein distance} \cite{Levenshtein1966} between 
the prediction and the target and the \emph{edit cost} of the action sequence.
Given input $\bv{x}^{(l)}$ with target $\bv{y}^{(l)}$, the loss from producing
an action sequence $\bv{a}$ is: 
\vspace{-5mm}
\begin{adjustwidth}{-0.25cm}{}%
\begin{equation}
\loss(\bv{a}, \bv{x}^{(l)}, \bv{y}^{(l)}) = \beta \, \textrm{distance}(\bv{y}, \bv{y}^{(l)}) + \textrm{cost}(\bv{a}),
\label{eq:seqlevloss}
\end{equation}%
\end{adjustwidth}%
where $\bv{y}$ is computed from $\bv{a}$ and $\bv{x}^{(l)}$ and
$\beta \geq 1$ is some penalty for unit distance.\footnote{We use unit costs    
to compute edit cost and distance.}                                             
The first term represents the task objective. The second term enforces that
the task objective is reached with a minimum number of edits.

The second term is crucial as it takes over the role of the character aligner.
Initially, we also experimented with only Levenshtein distance as loss, similar to previous
work on character-level problems \cite{Leblondetal2018,Bahdanauetal2017}. However,
models did not learn much, which we attribute to sparse training signal as
all action sequences producing the same $\bv{y}$ would incur the same
sequence-level loss, including intuitively very wasteful ones, e.g. first
deleting all of $\bv{x}^{(l)}$ and then inserting of all of $\bv{y}^{(l)}$.

\paragraph{Expert} The expert policy keeps track of the prefix of the target
$\bv{y}^{(l)}$ in the predicted sequence $\bv{y}_{<t}$ and returns actions that
lead to the completion of the suffix of $\bv{y}^{(l)}$ using an action sequence
with the lowest edit cost. The resulting prediction $\bv{y}$ attains the minimum
edit distance from $\bv{y}^{(l)}$. 
For example, if $\bv{x}^{(l)}=walk$ and $\bv{y}^{(l)}=walked$, the top of the
buffer is $\bv{h}_{3}$ representing $x_3=l$, and $\bv{y}_{<3} = wad$ due to a
sampling error from a roll-in with the model, the expert returns $\{ \COPY \}$.

\paragraph{Action-level loss} Given sequence-level losses, we compute the regret
for each action $a$:
\vspace{-5mm}
\begin{adjustwidth}{-0.2cm}{}%
\begin{equation}
r_t(a) = \loss(\bv{a}, \bv{x}^{(l)}, \bv{y}^{(l)}) -\!\!\!\! \min_{a^{\prime} \in A(\bv{s}_t)} \!\! \loss(\bv{a}^{\prime}, \bv{x}^{(l)}, \bv{y}^{(l)}),
\end{equation}%
\end{adjustwidth}%
where $\bv{a}$ (or $\bv{a}^{\prime}$) is the action sequence resulting from taking
$a$ (or $a^{\prime}$) at $\bv{s}_t$ and $A(\bv{s}_t)$ is the set of valid actions.
Thus, $r_t(a)$, which quantifies how much we
suffer from taking action $a$ relative to the optimal action under the current
policy, constitutes the direct blame of $a$ in the sequence-level loss.         

Classic IL employs cost-sensitive classification, with regrets making up costs
\cite{Daumeetal2009,Changetal2015}. Our initial experiments with cost-sensitive
classification resulted in rather inefficient training and not very effective
models. Instead, we choose to minimize the negative marginal log-likelihood of
all optimal actions \cite{Riezleretal2000,Goldberg2013,Ballesterosetal2016}.
Given the training data $T = \{(\bv{x}^{(l)}, \bv{y}^{(l)})\}_{l=1}^N$, the action-level loss is:
\vspace{-5mm}
\begin{adjustwidth}{-0.32cm}{}%
\begin{equation}
\Loss(T, \Theta)\! =\! - \sum^N_{l=1} \sum^m_{t=1} \log \! \sum_{a \in A_t}\! P(a \! \mid \! \bv{a}_{<t}, \bv{x}^{(l)}\!, \! \Theta),\!\!
\end{equation}%
\end{adjustwidth}%
where $A_t = \{a \in A(\bv{s}_t): r_t(a) = 0 \}$, the set of optimal actions
under the current policy. Depending on the roll-in schedule,  the next edit
$a_{t+1}$ is sampled either uniformly at random from $A_t$ or from the distribution of valid edits.
To include all the computed regrets into the loss, we also experiment with the
cost-augmented version of this objective \cite{Gimpel&Smith2010}, where regrets
function as costs.

The downside of IL is that roll-outs are costly. We avoid computing
most of the roll-outs by checking if an action increases the edit distance from
$\bv{y}^{(l)}$. If it does, we heuristically assign this action a regret of                   
$\beta$. We use this heuristic in both expert and model roll-outs.              

\section{Experiments}

\begin{table*}[h]
\begin{minipage}[t]{0.745\textwidth}\vspace{0pt}
\begin{flushleft}
\resizebox{1\linewidth}{!}{
\begin{tabular}{l|cccccccccc|c}
\bf Model       & \bf RU & \bf DE & \bf ES & \bf KA & \bf FI & \bf TR & \bf HU & \bf NV & \bf AR & \bf MT &\bf Avg.\\ \Xhline{2.5\arrayrulewidth} 

\textsc{med}     & 91.5  &  95.8  &  98.8  &  98.5  &  95.5  &  98.9  &  96.8  &  91.5  & \bf  99.3  &  89.0  &  95.6 \\
\textsc{soft}    & 92.2  &  96.5  &  98.9  &\bf 98.9  &  97.0  & \bf 99.4  &  97.0  & \bf 95.4  & \bf 99.3  &  88.9  & \bf 96.3 \\
\textsc{ha}      & 92.2  &  96.6  &  98.9  &  98.1  &  95.9  &  98.0  &  96.2  &  93.0  &  98.8  &  88.3  &  95.6 \\
\textsc{ha*}     & 92.0  & 96.3  & 98.9  & 97.9  & 95.8  & 97.6  & 98.8  & 92.1  & 95.1  & 87.8  & 95.2  \\
\textsc{ca}      & 91.9  & 96.4  & 98.8  & 98.3  & 96.5  & 97.7  & 98.9  & 92.1  & 94.6  & 87.7  & 95.3  \\
\hline\hline
\textsc{ca-d}    &\bf 92.4  &\bf 96.6  &\bf 98.9  & 98.7  & 97.2  & 98.5  & 99.3  & 95.2  & 96.5  &\bf 89.2  & 96.2  \\
\textsc{ca-r}    & 92.3  & 96.5  & 98.9  &\bf 98.9  &\bf 97.3  & 98.9  &\bf 99.4  & 95.2  & 96.1  & 88.8  & 96.2  \\
\end{tabular}}
\vspace{-2mm}
\captionof{table}{Results on \textsc{Sigmorphon 2016} data.\footnotemark}
\label{tab:sig2016}

\vspace{-2mm}

\begin{minipage}[t]{0.491\linewidth}\vspace{0pt}
\centering
\resizebox{1\textwidth}{!}{
\begin{tabular}{l|cccc|c}
\bf Model          &\bf \scalebox{0.7}{13SIA}   &\bf \scalebox{0.7}{2PIE}    &\bf  \scalebox{0.7}{2PKE}   &\bf \scalebox{0.8}{rP}   &\bf  Avg.  \\ \Xhline{2.5\arrayrulewidth}
\textsc{lat}       &   \bf 87.5 &    93.4    &    87.4    &   84.9   &  88.3    \\  
\textsc{nwfst}     &   85.1     &    94.4    &    85.5    &   83.0   &  87.0    \\
\textsc{ha$^*$}    &   84.6     &    93.9    &    88.1    &   85.1   &  87.9    \\
\textsc{ca}        &   85.0     &    94.5    &    88.0    &   84.9   &  88.1    \\
\scalebox{0.78}{\textsc{ha$^*$-mrt}} &   84.8      &   94.0     &   88.1     &   85.2   & 88.0 \\
\scalebox{0.9}{\textsc{ca-mrt}}   &   85.6      &\bf 94.6    &   88.0     &   85.3   & 88.4\\     
\hline\hline
\textsc{ca-d}     &   85.7     &   94.4     &\bf 88.4    &   85.1     &   88.4  \\
\textsc{ca-r}     &   85.6     &   94.4     &   88.3     &\bf 85.3    &\bf 88.4  \\
\textsc{ca-rm}    &   84.9     &   94.1     &   88.3     &   85.0     &   88.1  \\
\end{tabular}}
\end{minipage}\enskip
\begin{minipage}[t]{0.491\linewidth}\vspace{0pt}
\begin{center}
\resizebox{1\linewidth}{!}{  %
\begin{tabular}{l|cccc|c}
\bf Model       & \bf EU & \bf EN & \bf GA & \bf TL &\bf Avg.\\ \Xhline{2.5\arrayrulewidth}  
\textsc{lat}    & 93.6       & 96.9        & 97.9      & 88.6        & 94.2   \\
\textsc{nwfst}  & 91.5       & 94.5        & 97.9      & 97.4        & 95.3   \\
\textsc{lem}\footnotemark    & 96.5       & 96.3        & \bf 98.7  & \bf 98.8    & 97.6  \\
\textsc{ha$^*$} & \bf 97.0   & \bf 97.5    & 97.9      & 98.3        & \bf 97.7   \\    
\textsc{ca}     & 96.3       &  96.9       &  97.7     &  98.3       &  97.3   \\                 
\hline\hline
\textsc{ca-d}  &  96.1       &  97.0       &  97.7     &  98.4       &  97.3   \\
\textsc{ca-r}  &  96.6       &  97.2       &  97.5     &  98.3       &  97.4   \\
\textsc{ca-rm} &  96.5       &  97.0       &  97.8     &  98.3       &  97.4   \\
\end{tabular}}
\end{center}
\end{minipage}

\vspace{-2mm}
\begin{minipage}[t]{0.51\linewidth}\vspace{0pt}
\captionof{table}{Results on \textsc{celex} data.}
\label{tab:celex}
\end{minipage}\enskip
\begin{minipage}[t]{0.47\linewidth}\vspace{0pt}
\captionof{table}{Lemmatization results.}
\label{tab:lem}
\end{minipage}

\end{flushleft}
\end{minipage}\enskip
\begin{minipage}[t]{0.24\linewidth}\vspace{0pt}
\begin{minipage}[t]{\linewidth}\vspace{0pt}
\begin{flushright}
\resizebox{1\linewidth}{!}{  %
\begin{tabular}{l|cc}  
\bf Model           & \bf L & \bf M  \\ \Xhline{2.5\arrayrulewidth} 
\scalebox{0.9}{\textsc{sgm17top}} & 50.6 & 82.8 \\   
\textsc{ha$^*$}     & 31.5 & 80.2 \\  
\textsc{ca}         & 48.8 & 81.0 \\  
\textsc{ha$^*$-mrt} & 33.1 & 81.5 \\  
\textsc{ca-mrt}     & 49.9 & 82.9 \\  
\hline\hline
\scalebox{0.9}{\textsc{ca-mrt-a}} & 49.9 & 82.7 \\  
\textsc{ca-d}      & 50.3 & 82.6 \\
\textsc{ca-r}      & \bf 51.6 & 83.8 \\
\textsc{ca-rm}     & 50.6 &\bf 84.0 \\
\end{tabular}%
}
\captionsetup{font=small,labelfont=small}
\captionof{table}{Results on \textbf{L}ow and \textbf{M}edium settings of \textsc{Sigmorphon 2017} data (averaged over 52 languages).}
\label{tab:sig2017}
\end{flushright}
\end{minipage}

\smallskip

\begin{minipage}[t]{\linewidth}\vspace{0pt}
\footnotesize{
\textbf{\textsc{-mrt}}: minimum risk training;
\textbf{\textsc{-mrt-a}}: MRT with action cost in the loss;
\textbf{\textsc{-d}}: only expert roll-outs; 
\textbf{\textsc{-r}}: expert and model roll-outs;
\mbox{\textbf{\textsc{-rm}}}: softmax-margin, expert and model roll-outs
}
\end{minipage}
\end{minipage}

\medskip

\begin{minipage}[t]{1\linewidth}
\textbf{Experimental results.} Soft-attention seq2seq models:
\textsc{med}=\newcite{Kann&Schutze2016b} (cited from \newcite{Aharoni&Goldberg2017}), 
\textsc{soft}=\newcite{Aharoni&Goldberg2017}, 
\textsc{lem}=\newcite{Bergmanis&Goldwater2018}.
WFSTs: \textsc{lat}=\newcite{Dreyeretal2008},
\textsc{nwfst}=\newcite{Rastogietal2016}.
Transition-based models:
\textsc{ha}=\newcite{Aharoni&Goldberg2017}, 
\textsc{sgm17top}=\newcite{Makarovetal2017}, 
and from \newcite{Makarov&Clematide2018}:
\textsc{ha$^*$}=reimplementation of \textsc{ha}, 
\textsc{ca}=model in \S\ref{sec:model}, and \textsc{ha$^*$-mrt}, \textsc{ca-mrt}
(risk=normalized edit distance).
We report exact-match accuracies for ensembles of 5 models 
(\textsc{Sigm. 2016} and \textsc{2017}) and single-model averages over
5 folds (\textsc{Celex}) and 10 folds (lemmatization).
\end{minipage}
\end{table*}

We demonstrate the effectiveness of our approach on three tasks: inflection
generation (using the typologically diverse \textsc{Sigmorphon 2016} and
\textsc{Sigmorphon 2017} datasets of \newcite{Cotterelletal2016,Cotterelletal2017}),
reinflection (the small-sized German \textsc{Celex} dataset of
\newcite{Dreyeretal2008}), and lemmatization (the standard subset of the
\newcite{Wicentowski2002} dataset). 

We use character and feature embeddings of size 100 and 20, respectively, and
one-layer LSTMs with hidden-state size 200. Following
\citeauthor{Aharoni&Goldberg2017}, for every character
$c \in \Sigma_x \cap \Sigma_y$, we let $A($\INSERT[c]$) := E(c)$, i.e.
the same embedding represents both $c$ and the insertion of $c$. We
optimize with ADADELTA \cite{Zeiler2012}, use early stopping and batches of
size 1. We set the penalty for unit distance $\beta=5$ 
and roll in with an inverse-sigmoid decay schedule as in \newcite{Bengioetal2015}. 
\textsc{ca-d} models are trained with expert roll-outs only (as is often the
case in dynamic-oracle parsing). \textsc{ca-r} and \textsc{ca-rm} models mix
expert and learned roll-outs with probability 0.5 as in \newcite{Changetal2015}.
\textsc{ca-rm} models optimize softmax-margin.

For comparison, we also train models with MRT (\textsc{ca-mrt-a})
as in \newcite{Shenetal2016}, using a global objective similar to our sequence-level
loss (Eq.~\ref{eq:seqlevloss}). We use batches of at most 20 unique samples
per training example. 
The risk is a                                          
convex combination of normalized Levenshtein distance and the action sequence
cost, which we min-max scale, within a batch, to the [0, 1] interval.

We decode all our models using beam search with beam width 4.

\addtocounter{footnote}{-2}
\stepcounter{footnote}\footnotetext{
Language codes: RU=Russian, DE=German, ES=Spanish,
KA=Georgian, FI=Finnish, TR=Turkish, HU=Hungarian, NV=Navajo, AR=Arabic,
MT=Maltese, EU=Basque, EN=English, GA=Irish, TL=Tagalog.
}

\stepcounter{footnote}\footnotetext{Personal communication.}

Our approach performs best on most languages of the
\textsc{Sigmorphon 2016} data (Table~\ref{tab:sig2016}) 
and both limited-resource settings of \textsc{Sigmorphon 2017} (Table~\ref{tab:sig2017}).
It achieves marginal improvement over an MRT model on the reinflection task
(Table~\ref{tab:celex}) with consistent gains on the 2PKE$\mapsto$z transformation
\cite{Dreyeretal2008}, that involves infixation.
Using mixed roll-outs (\textsc{ca-r}, \textsc{ca-rm}) improves performance on the \textsc{Sigmorphon 2017}  
inflection data (Table~\ref{tab:sig2017}), otherwise the results are close to
\mbox{\textsc{ca-d}}.  We also note strong gains over
\textsc{ca-mrt-a} trained with a similar global loss (Table~\ref{tab:sig2017}).
Generally, improvements are most pronounced in inflection generation, the only
task where the model could profit from adjusting alignment to available
feature information (cf. Table~\ref{tab:lem}).

We take a closer look at the results in the \textsc{Sigmorphon 2017} medium
data-size setting (1,000 training examples per language). 
\textsc{ca-rm}
makes the largest performance gains on languages with complex morphological phenomena
(Semitic and Uralic languages, Navajo) and an above average number
of unique morpho-syntactic descriptions. Khaling and
Basque, outliers with 367 and 740 unique morpho-syntactic descriptions in the
training data, are among the top five languages with the largest gains. The lowest gains
and rare losses are made for Romance and Germanic languages and languages with
many unique morpho-syntactic descriptions but regular morphologies (Quechua, Urdu/Hindi).

\vspace{3mm}
\section{Related work}
Traditionally, morphological string transduction has been approached with
discriminative weighted finite-state transducers
\cite{Rastogietal2016,Cotterelletal2014,Dreyeretal2008,Eisner2002}. 
\newcite{Yuetal2016} and \newcite{Graves2012} tackle the modeling of unbounded
dependencies in the output, while preserving latent monotonic hard character
alignment.
\newcite{Faruquietal2016,Kann&Schutze2016b} successfully apply
seq2seq modeling to the task. \newcite{Aharoni&Goldberg2017} introduce a neural
version of the transition-based model over edits.
\newcite{Makarov&Clematide2018} show gains from using the copy edit and address
the MLE training biases with MRT.

The limitations of teacher forcing have recently been the focus of intense research
\cite{Edunovetal2018,Wiseman&Rush2016,Shenetal2016}, including 
the adaptation of RL methods \cite{Ranzato16,Bahdanauetal2017}.
Most of these approaches require warm start with an MLE model and               
themselves introduce discrepancies between training with sampling and
search-based decoding. Such biases do not arise from IL, which
has recently been proposed for seq2seq models \cite{Leblondetal2018}. Our
approach, related to \newcite{Leblondetal2018}, additionally addresses
the problem of spurious ambiguity, which is not present in seq2seq models.





\section{Conclusion}
We show that training to imitate a simple expert policy results in an effective
neural transition-based model for morphological string transduction. The
fully end-to-end approach addresses various shortcomings of previous training
regimes (the need for an external character aligner, warm-start initialization,
and MLE training biases), and leads to strong empirical results.  
We make our code and predictions publicly available.\footnote{\href{https://github.com/ZurichNLP/emnlp2018-imitation-learning-for-neural-morphology}{{https://github.com/ZurichNLP/emnlp2018-imitation-learning-for-neural-morphology}}}

\section*{Acknowledgments}
We thank Mathias M\"{u}ller for help with the manuscript, Toms Bergmanis for
sharing with us the results of his system, and the reviewers for interesting and
helpful comments. We also thank the anonymous COLING reviewer who suggested that
we should look at spurious ambiguity. Peter Makarov has been
supported by European Research Council Grant No.~338875, Simon Clematide by the
Swiss National Science Foundation under Grant \mbox{No.~CRSII5\_173719}. 

\bibliography{morph}
\bibliographystyle{acl_natbib_nourl}

\end{document}